\documentclass[10pt, a4paper]{article}

\usepackage[]{lrec-coling2024} 

\newcommand{\bluefont}[1]{ {\color{blue}{#1}}}
\newcommand{\redfont}[1]{{\textcolor{red}{#1}}}

\usepackage{bbm}
\usepackage{multirow}
\usepackage{amsmath}
\usepackage{amssymb}
\usepackage{booktabs}
\usepackage{siunitx}
\usepackage{tabularx}
\usepackage{array}
\usepackage{xcolor}

\title{FaiMA: \underline{F}eature-\underline{a}ware \underline{I}n-context Learning for \underline{M}ulti-domain \underline{A}spect-based Sentiment Analysis}

\name{
  \begin{tabular}{c}
    Songhua Yang$^1$, Xinke Jiang$^2$\sthanks{~~Songhua Yang and Xinke Jiang contributed equally to this research.}~, Hanjie Zhao$^1$\\
    Wenxuan Zeng$^2$, Hongde Liu$^1$, Yuxiang Jia$^1$\sthanks{~~Yuxiang Jia is the corresponding author.}\\
  \end{tabular}
}

\address{$^1$~Zhengzhou University, Henan, China \\
         $^2$~Peking University, Beijing, China \\
         \{suprit,thinkerjiang\}@foxmail.com, hjzhao\_zzu@163.com\\
         zwx.andy@stu.pku.edu.cn,
         lhd\_1013@gs.zzu.edu.cn,
         ieyxjia@zzu.edu.cn
         }

\abstract{
Multi-domain aspect-based sentiment analysis (ABSA) seeks to capture fine-grained sentiment across diverse domains. While existing research narrowly focuses on single-domain applications constrained by methodological limitations and data scarcity, the reality is that sentiment naturally traverses multiple domains. Although large language models (LLMs) offer a promising solution for ABSA, it is difficult to integrate effectively with established techniques, including graph-based models and linguistics, because modifying their internal architecture is not easy. To alleviate this problem, we propose a novel framework, \textbf{F}eature-\textbf{a}ware \textbf{I}n-context Learning for \textbf{M}ulti-domain \textbf{A}BSA (FaiMA). 
The core insight of FaiMA is to utilize in-context learning (ICL) as a feature-aware mechanism that facilitates adaptive learning in multi-domain ABSA tasks. 
Specifically, we employ a multi-head graph attention network as a text encoder optimized by heuristic rules for linguistic, domain, and sentiment features.
Through contrastive learning, we optimize sentence representations by focusing on these diverse features. Additionally, we construct an efficient indexing mechanism, allowing FaiMA to stably retrieve highly relevant examples across multiple dimensions for any given input. To evaluate the efficacy of FaiMA, we build the first multi-domain ABSA benchmark dataset. Extensive experimental results demonstrate that FaiMA achieves significant performance improvements in multiple domains compared to baselines, increasing F1 by 2.07\% on average. Source code and data sets are available at \url{https://github.com/SupritYoung/FaiMA}.
 \\ \newline 
 \Keywords{Multi-domain Aspect-based Sentiment Analysis, Graph Neural Networks, Large Language Model, In-Context Learning, Linguistics} 
}

\begin{document}

\maketitleabstract

\section{Introduction}

In the highly interconnected digital era, a myriad of social media platforms are continually emerging \citep{roccabruna2022multi}. These platforms generate a vast corpus of user reviews across various domains, providing a rich reservoir of sentiment-related information. 
For years, aspect-based sentiment analysis (ABSA) has emerged as a long-standing solution to this problem \citep{Pang2008OpinionMA, Zhang2012SentimentAA, Schouten_Frasincar_2016}. 
ABSA is a fine-grained sentiment analysis task that can meticulously extract the sentiment polarity of users towards specific aspects.
However, the majority existing ABSA methods are confined to single-domain applications, struggling to capture the multifaceted sentiment information prevalent in the real world. Traditional approaches often encounter generalization challenges across multiple domains, limiting the practical and broad-scale applicability of ABSA \citep{Luo2022ChallengesFO}. 
Customizing models and annotating data for each domain is inefficient and costly, especially in resource-limited settings.

\begin{figure}[htbp]
  \centering
\includegraphics[width=0.45\textwidth]{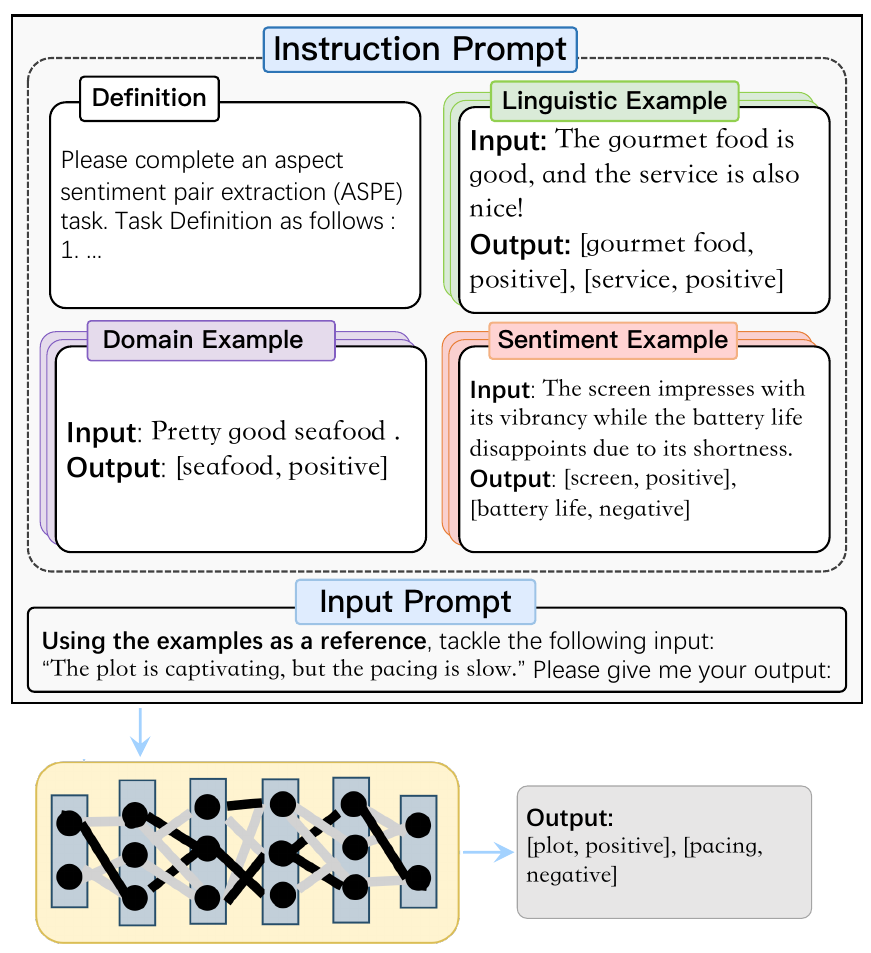}
  \caption{An example of feature-aware in-context learning for ABSA. By selecting one relevant example on each of the three features, sufficient reference is provided for LLM.}
  \label{fig:motivation}
\end{figure}

Fortunately, the advent of Large Language Models (LLMs) can imbue the multi-domain ABSA with renewed optimism, owing to their remarkable generalization and cross-domain capabilities \citep{Wang2023IsCA, Zhang2023SentimentAI}. Trained on extensive, multi-domain corpora, LLMs assimilate a broad spectrum of common sense and domain-agnostic knowledge, equipping them with the ability to discern nuanced differences and linguistic subtleties across various domains \citep{Zhao2023ASO, dillion2023can, Yang_Zhao_Zhu_Zhou_Xu_Jia_Zan_2023, luo2024kuaiji}. Moreover, emerging in-context learning (ICL) techniques demonstrate that task-specific performance can be significantly amplified by simply incorporating concise, task-relevant instructions, demonstrations, and examples into the prompts \citep{Ye2023InContextIL,jiang2023think}. Although initial research has begun to probe the potential of LLMs and associated techniques in ABSA \citep{Fei2023ReasoningIS, Scaria2023InstructABSAIL, Varia2022InstructionTF}, empirical investigations explicitly focusing on multi-domain ABSA are still notably scarce.

Another line of ABSA research focuses on graph neural networks (GNNs) and linguistic features \citep{Chen2022EnhancedMG}. Linguistic knowledge, epitomized by syntactic and part of speech, is widely regarded as essential for solving ABSA tasks, as they share intricate connections with the relationships between sentiment elements \citep{zhang2022survey, nazir2020issues}. Numerous studies demonstrated that leveraging these linguistic features to construct relationships between words and leveraging the unique message-passing mechanism of GNNs can effectively capture complex and latent relationships among sentiment elements \citep{Wu2021LearnFS, chen2021semantic, yang2023improving, shi2023syntax, Zhong_Ding_Liu_Du_Jin_Tao_2023}. Features such as domain and sentiment structure are also valuable in multi-domain ABSA \citep{Wu2020GridTS, Gong2020UnifiedFA}. In the context of multi-domain ABSA's complex and diverse landscape, general linguistic features can provide substantive support, while specific domain information can serve as unique augmentative features.
On the other hand, LLMs are often perceived as inscrutable "black box", making it challenging to directly modify their internal architecture or incorporate additional features \citep{luo2023augmented, Zhao2023ASO}. Solely fine-tuning LLMs for ABSA fails to integrate the wealth of domain-specific expertise and the intrinsic relationships between parts of speech and syntax. Seamlessly integrating these well-established traditional methods with cutting-edge LLMs to fully unleash their collective potential remains a pivotal challenge in current research.

Incorporating semantically similar examples into the instructions can significantly enhance the performance of LLMs on specific tasks \citep{liu2021makes}. Unlike unsupervised strategies, supervised example retrieval methods have proven to be more effective \citep{rubin2022learning, Zhang2022ActiveES}. In light of this, we propose the following critical hypothesis: \textbf{ICL is not only a tool to guide the model but also an efficient feature-aware mechanism}. We further hypothesize that the stable retrieval of representative examples for various features, followed by their precise incorporation into fine-tuning instructions, can give the model a structured and enriched feature context, as shown in \ref{fig:motivation}. This, in turn, substantially enhances its performance on the target task. By undergoing supervised fine-tuning (SFT) on extensive data, LLMs with strong comprehension capabilities can fully grasp, understand, and apply these features, achieving marked performance improvements in ABSA tasks.

In light of the above, we introduce a novel \textbf{F}eature-\textbf{a}ware \textbf{I}n-context Learning for \textbf{M}ulti-Domain \textbf{A}BSA (FaiMA) framework. FaiMA ingeniously amalgamates traditional techniques with cutting-edge LLMs, using ICL as the linchpin that coherently integrates these components. Explicitly, we architect a Multi-head Graph Attention Network Encoder (MGATE) to function as the sentence encoder. Employing a multi-headed Graph Attention Network (GAT) architecture, MGATE concentrates on a panoply of linguistic, domain, and sentiment features, thereby engendering a unique sentence encoding paradigm. 
The essence of MGATE is its ability to wisely choose examples that are highly aligned with any given input across a variety of feature dimensions.
To achieve this, the ICL technique is combined with SFT in the training stage to impart LLMs a nuanced, feature-aware understanding and learning capacity.
In order to make it easier to retrieve the most relevant examples, we craft a set of heuristic rules that quantify sentence similarity across various feature dimensions. 
This approach generates a balanced mix of positive and negative samples for the next MGATE contrastive learning training. After training and optimizing sentence representations, the MGATE can achieve a refined understanding of the features critical for multi-domain ABSA tasks, producing high-quality sentence representations. Building on this, we select the most similar examples across features and insert them into instruction prompts during both the training and inference stages, further enhancing performance for multi-domain ABSA tasks.

In response to the lack of specialized multi-domain ABSA datasets, we also constructed a benchmark dataset named MD-ASPE, which combines 16,000 sentences across nine diverse domains. Extensive experiments show that FaiMA performs in all these domains and increases average performance by 2.07\% compared to baseline models.

Our contributions can be summarized as follows:
\begin{itemize}
    \item We introduce FaiMA, a novel framework based on LLMs for multi-domain ABSA tasks, demonstrating that ICL can be an effective feature-aware tool.
    \item We propose a sentence encoding model, MGATE, which combines multi-head GAT and contrastive learning. It fully integrates linguistic, domain, and sentiment features, allowing the robust retrieval of highly relevant examples in multiple dimensions.
    \item We present MD-ASPE, the first benchmark dataset for multi-domain ABSA. Extensive experiments demonstrate that our method achieves state-of-the-art performance across nearly all domains and on average.
\end{itemize}

\section{Related Work}

Historically, extensive research has demonstrated the universal applicability of specific features for ABSA tasks \cite{zhang2022survey}. 
 For example, dependency parse trees and part-of-speech tagging naturally captured relationships between words and were considered crucial linguistic features for tackling ABSA; they were closely related to underlying sentiment elements \citep{Zhang2019SyntaxAwareAS, Wu2021LearnFS, chen2021semantic, shi2023syntax}.
 Furthermore, \citet{Wu2020GridTS, Chen2022EnhancedMG} introduced a grid tagging scheme, formalizing ABSA as a task to predict the types of edge relations between words. Given that ABSA spans multiple domains, domain-specific information is often considered a crucial feature \citep{Gong2020UnifiedFA, Tian2021CrossDomainEA}. Since ABSA can be considered an edge-sensitive task, GNN-based models demonstrated remarkable performance \citep{Zhang2019SyntaxAwareAS, Wang2022AspectbasedSA, Zhang2022SSEGCNSA}. 
 Significantly, the multi-head GAT model, which can flexibly focus on multiple features, achieved superior performance \citep{Wang2020RelationalGA, Liang2022BiSynGATBA, yang2023improving}.

Recently, LLMs like ChatGPT or LLaMA have achieved groundbreaking success \citep{Touvron2023LLaMAOA, Touvron2023Llama2O}. With the increasing scale of LLMs, novel techniques such as ICL \citep{Ye2023InContextIL} and Chain of Thought (CoT) \citep{Wei2022ChainOT} emerged. ICL demonstrates that adding detailed instructions and examples to task prompts can significantly enhance task performance, whether in zero-shot inference or supervised training. Current research has begun to investigate optimal example selection to further augment ICL's capabilities. Studies \citep{liu2021makes, 2022Rethinking} found that choosing examples semantically and label-wise closer to the actual input is more effective. Moreover, \citet{rubin2022learning, Zhang2022ActiveES} revealed that training a retriever in a supervised way to find more relevant examples is a more practical approach.

\begin{figure*}[htbp]
  \centering
  \includegraphics[width=\textwidth]{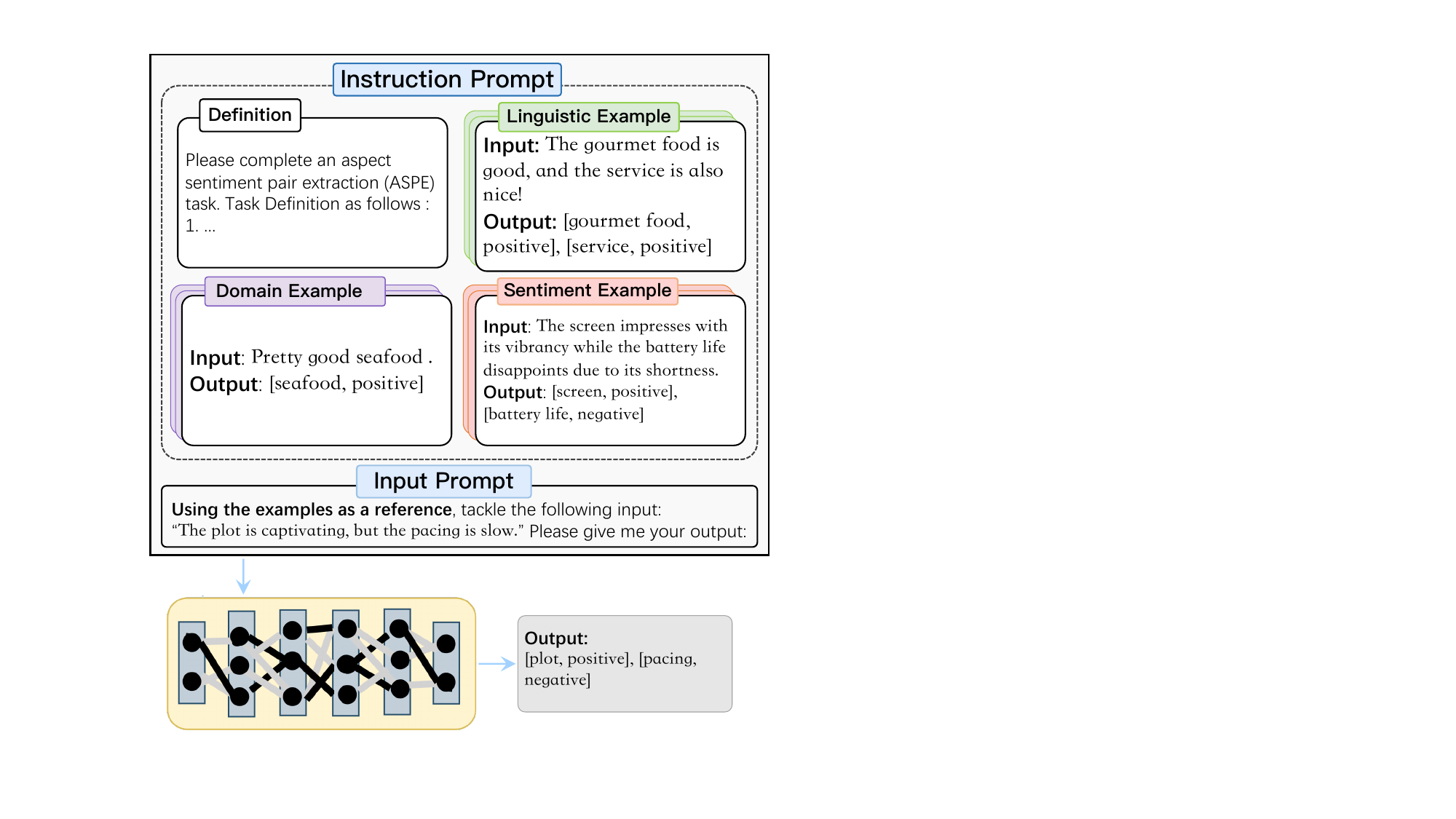}
  \caption{The overall architecture of FaiMA: MGATE training part and example retrieval part. MGATE training involves three steps: heuristic rules for positive/negative pairs generation, multi-head graph attention network to embed sentences upon three features,  
  and contrastive learning. The diagram to the far right illustrates an ICL process that reliably fetches three domain-relevant and global average samples for any input sentence.}
  \label{fig:framework}
\end{figure*}

Traditional methods based on Small Language Models (SLMs) for multi-domain ABSA showed limited performance \citep{Hu2019OpenDomainTS, Ji2020AUL, Luo2022ChallengesFO}. Previous approaches usually trained models for every domain, leading to computational and resource costs. Recent work has begun to explore the combination of LLMs for ABSA. For example, \citet{Varia2022InstructionTF} demonstrated a few-shot generalizability across various ABSA subtasks using SFT and multi-task learning. \citet{Scaria2023InstructABSAIL} employed ICL with fixed examples to achieve marked performance, while \citet{Fei2023ReasoningIS} designed a multi-turn CoT for understanding implicit sentiments and opinions. These studies provide initial evidence of the substantial research potential of LLM in ABSA tasks.

\section{Methodology}

In this section, we introduce our proposed FaiMA framework, depicted in Figure \ref{fig:framework}.
In \S\ref{MGATE}, we describe the MGATE, elaborate on the heuristic rules, and contrastive learning. 
In \S\ref{Retriver}, we discuss how to perform feature-aware example retrieval, along with the specific implementation of the ICL strategy.

\subsection{Problem Definition}

As a fine-grained sentiment analysis task, ABSA can be formalized as a hybrid task of extraction and classification. Given a sentence \( L = \{t_1, t_2, ..., t_n\} \), where \( t_i \) represents the \( i \)-th word in this sentence, and the multi-domain ABSA refers to the extraction of all sentiment pairs \(P = \{(A_1, S_1), ..., (A_m, S_m)\}\) in $L$ jointly for different domains. \footnote{This task is also referred to as aspect sentiment pair extraction (ASPE) in some literature.}
Formally, $A$ indicates an entity or phase in the sentence \(S\) that are related to sentiment, defined as \(A = \{a_1, a_2, ..., a_p\} \subseteq L \) and the sentiment polarity $ \mathcal{S} \in \{\text{positive}, \text{negative}, \text{neutral}\}$.

\subsection{Multi-head Graph Attention Network Encoder}
\label{MGATE}
For a long time, the linguistic, domain, sentiment features, and the GNN model have been crucial components for ABSA \cite{zhang2022survey, Chen2022EnhancedMG}.
In light of this, we propose MGATE, a submodel designed to investigate and understand the intricate interplay of these three complex features between words within sentences.
To enhance the training process, we develop a set of sophisticated heuristic rules to generate positive and negative training sentence pairs for each feature, and then employ contrastive learning to train the graph neural encoder, optimizing sentence representations from these three perspectives. The detailed implementation is as follows.

\subsubsection{Feature Selection and Heuristic Rules}
\label{heuristic}

To accommodate the properties of the three different features, we devise unique processing rules for each.
For linguistic and sentiment features, direct conversion to trainable positive and negative sample pairs presents challenges. Therefore, we design a set of heuristic algorithms to precisely calculate the similarity between two sentences given in the ABSA task.

\textbf{Linguistic Similarity} Linguistic knowledge has always been considered an essential resource to solve the ABSA task \cite{chen2021semantic, shi2023syntax}. We select the most representative part-of-speech combinations and syntactic dependency types and define refined feature modeling methods. Initially, for a sentence $L$, using a parser to establish part-of-speech combination matrices $R^{pos} \in \mathbb{R}^{n \times n}$ and syntactic dependency matrices $R^{dep} \in \mathbb{R}^{n \times n}$, where each type of relationship corresponds to a unique numerical ID.

In the ABSA task, the aspect is always considered the key to solving this task and is closely related to other elements, so we redefine the aspects and opinions in the sentence as central words \(C = A = \{c_1, c_2, ...,c_k\} \). 
The central word for multi-token phrases is selected based on its highest number of relationships with other words.
For each main word \(c_k\), we assign weights to the other words in the sentence using a Gaussian function, ensuring that terms closer to the main word receive more significant weight. It is defined as:
\begin{equation}
    W(c_k) = \left[ e^{-\frac{(1 - k)^2}{2\sigma^2}}, ..., e^{-\frac{(n - k)^2}{2\sigma^2}} \right] \in {\mathbb{R}^{n} }.
\end{equation}

The similarity calculation uses the weighted Hamming distance \(\texttt{HM}(\cdot)\), which can effectively capture minor structural changes in the sentence and amplify the influence of core words nearby, considering comprehensive linguistic structures beyond direct word-to-word connections, defined as the weighted Hamming distance:
\begin{multline}
    H(i, j) = W(c_i) \circ \text{HM}([R^{dep}(c_i), R^{pos}(c_i)], \\
    [R^{dep}(c_j), R^{pos}(c_j)]),
\end{multline}
where \( \circ \) denotes the dot product operation, and \(c_i \in C_1 \) and \(c_j \in C_2 \) are the central word sets of two sentences \( L_1 \) and \( L_2 \), respectively. The overall similarity distance is calculated as:
\begin{equation}
    D(L_1, L_2) = \frac{1}{|C_1||C_2|} \sum_{i=1}^{|C_1|} \sum_{j=i}^{|C_2|} H(i, j).
\end{equation}

Finally, the linguistic similarity score between the two sentences is obtained through the 
\(\texttt{Sigmoid}(\cdot)\) function:
\begin{equation}
S_{Lig}(L_1, L_2) = \texttt{Mean}(\texttt{Sigmoid}(D(L_1, L_2)))
\end{equation}
where $\texttt{Mean}$ refers to the averaging operation, $D(L_1, L_2) \in \mathbb{R}^{min(n,m)^2}$ and the final output is a scalar. This strategy combines part-of-speech, syntactic dependencies, and core word concepts, providing an effective quantitative measure of sentence linguistic similarity for the ABSA task.

\textbf{Domain Similarity} ~ In Multi-domain ABSA, texts from different domains may possess entirely different features and styles, while texts from the same domain share similar background knowledge and emotional objects. Therefore, taking into account domain similarity becomes a critical factor. We define a simple binary metric to measure this. Given two sentences \( L_1 \) and \( L_2 \) that belong to domains \( D_1 \) and \( D_2 \) respectively, the domain similarity \( S_{\text{Dom}} = \mathbbm{1}_{D1=D2} \), where \(\mathbbm{1}(\cdot)\) is the indicator function, taking the value of 1 if the condition is met and 0 otherwise.

\textbf{Sentiment Similarity} ~ Sentiment similarity in ABSA is not directly measurable. Review text often contains different sentiment polarities across multiple aspects, especially in long or complex sentences. To capture these nuanced variations, we introduce a sentiment vector representation. For each sentence \( L \), we define a sentiment vector \( \mathbf{v} = [n_{pos}, n_{neu}, n_{neg}] \), where \( n_{pos}, n_{neu}, n_{neg} \) represent the count of positive, neutral, and negative sentiments in the text, respectively. For two sentences \( L_1 \) and \( L_2 \) and their corresponding sentiment vectors \( \mathbf{v_1} \) and \( \mathbf{v_2} \), their sentiment similarity is calculated as follows:
\begin{multline}
    S_{sen}(L_1, L_2) = \frac{1}{2} \cdot \frac{\mathbf{v_1} \circ \mathbf{v_2}}{\|\mathbf{v_1}\| \|\mathbf{v_2}\|} + \frac{1}{2}.
\end{multline}

Here, \( \circ \) denotes the dot product between the two vectors, and \( \|\mathbf{v}\| \) represents the Euclidean norm of the vector.

Through the aforementioned method, we obtain a quantified inter-sentence similarity measure as follows:
\begin{equation}
S(L_1, L_2) = \left [S_{\texttt{Lig}}(L_1, L_2), S_{\texttt{Dom}}(L_1, L_2), S_{\texttt{Sen}}(L_1, L_2) \right]
\end{equation}
which integrates the three feature dimensions of linguistics, domain, and sentiment. By further setting three thresholds \( \theta_{\texttt{Lig}}, \theta_{\texttt{Dom}}, \theta_{\texttt{Sen}} \), these continuous similarity values can be mapped into a three-dimensional 0-1 tensor \( T \). \( T_{ijk} \) represents the value of sentence \( i \) and \( j \) in dimension \( k \). The tensor \( T = [\mathbbm{1}_{S_{\texttt{Lig}}\ge \theta_{\texttt{Lig}}}, \mathbbm{1}_{S_{\texttt{Dom}}\ge \theta_{\texttt{Dom}}}, \mathbbm{1}_{S_{\texttt{Sen}}\ge \theta_{\texttt{Sen}}}] \) serves as a multi-dimensional matrix of positive and negative samples, and will be subsequently used for contrastive learning to optimize the training of the graph encoder.

\begin{table*}[h]
  \centering
  \caption{The statistics of train and test datasets. \#S, \#P represent the number of sentences and sentiment pairs in the dataset, and \#Pos, \#Neg, \#Neu refer to the amount of corresponding sentiment polarity.} 
  \label{tab:data}\resizebox{0.85\linewidth}{!}{
  \begin{tabular}  {ccccccccccc}
    \toprule
    \multirow{2}{*}{Dataset}
     & \multicolumn{5}{c}{Train set} & \multicolumn{5}{c}{Test set} \\
     \cmidrule(lr){2-6} \cmidrule(lr){7-11}
     & \#S  & \#P  & \#Pos & \#Neg & \#Neu & \#S  & \#P  & \#Pos & \#Neg & \#Neu \\
    \midrule
    laptop     & 1148 & 1384 & 745  & 518  & 121  & 339  & 418  & 279  & 93   & 46   \\
    restaurant     & 1500 & 2125 & 1525 & 452  & 148  & 496  & 726  & 555  & 128  & 43   \\
    twitter & 1500 & 1500 & 353  & 390  & 757  & 500  & 500  & 134  & 112  & 254  \\
    books     & 1411 & 1780 & 1282 & 445  & 53   & 421  & 538  & 394  & 127  & 17   \\
    clothing  & 1303 & 1567 & 1158 & 381  & 28   & 318  & 369  & 274  & 88   & 7    \\
    device    & 948  & 1405 & 905  & 500  & 0    & 482  & 696  & 480  & 216  & 0    \\
    finance   & 1500 & 2139 & 675  & 608  & 856  & 500  & 593  & 291  & 220  & 82   \\
    hotel     & 1468 & 1963 & 1856 & 100  & 7    & 500  & 678  & 636  & 42   & 0    \\
    service   & 1432 & 1842 & 1032 & 698  & 112  & 500  & 618  & 350  & 229  & 39   \\
    \midrule
    \textbf{Overall}   & \textbf{12210}& \textbf{15705}& \textbf{9531} & \textbf{4092} & \textbf{2082} & \textbf{4056} & \textbf{5136} & \textbf{3393} & \textbf{1255} & \textbf{488}  \\
    \bottomrule
  \end{tabular}}
\end{table*}

\subsubsection{Multi-head Graph Attention Network}
\label{network}

The multi-head GAT is designed to discern intricate interrelations among linguistic, domain, and sentiment features at the token level. To this end, we deploy three distinct sub-linear layers that serve as encoders for token adjacency matrices, subsequently leveraging multi-head graph attention networks~\cite{zhang2024infinitehorizon,luo2024time} for aggregating features. The aggregated features are globally pooled to produce a graph-level (or sentence-level) representation.

Given a sentence \( L = \{w_1, w_2, \ldots, w_n\} \), we apply a pre-trained BERT model to encode word token:
\begin{equation}
(h_1, h_2, \ldots, h_n) =  \texttt{BERT}(w_1, w_2, \ldots, w_n),
\end{equation}
we denote sentence's token vectors $H=(w_1, w_2, \ldots, w_n)$. Then an Adaptive Adjacency Matrix is employed as the feature propagation matrix for these token vectors. For instance, the Adaptive Adjacency Matrix corresponding to linguistic features denoted as \( A^{{\text{(Lig)}}} \), is computed through:
\begin{equation}
A^{{\texttt{(Lig)}}} = \texttt{Sigmoid}(H W^{{\texttt{(Lig)}}} H^\mathsf{T}).
\end{equation}
and \( W^{\texttt{(Lig)}} \) is learnable weight to linguistic feature. We also compute \( A^{{\texttt{(Dom)}}} , A^{{\texttt{(Sen)}}} \) by learnable weights $W^{\texttt{(Dom)}}, W^{\texttt{(Sen)}}$ for domain and sentiment features, respectively.

The attention coefficients between the \(i^{th}\) and \(j^{th}\) tokens are then calculated as:
\begin{equation}
\alpha_{ij} = \frac{\exp(\texttt{LeakyReLU}(\vec{a}^\mathsf{T}[W_a h_{i} \| W_a h_{j}]))}{\sum_{A_{ik}^{\texttt{(Lig)}} > \delta}\exp(\texttt{LeakyReLU}(\vec{a}^\mathsf{T}[W_a h_{i} \| W_a h_{k}]))} 
\end{equation}
where \( \delta \) serves as a threshold to filter out noise in the adjacency matrix. $\vec{a}, W_a$ are learnable weights.

Then the token-level representation for the \(i^{th}\) token \( E_i \), is computed via attention mechanism:
\begin{equation}
E_{i} = \texttt{LayerNorm}\bigl(\sum_{A^{{\texttt{(Lig)}}}_{ij} > \delta} \alpha_{ij} A^{{\texttt{(Lig)}}}_{ij} W_a h_{j}\bigl).
\end{equation}

To synthesize the graph-level representation, an average pooling operation is applied across all token-level features:

\begin{equation}
E^{\texttt{(Lig)}} = \texttt{AVG}(E_1, \ldots, E_n).
\end{equation}

Lastly, we compute the comprehensive graph-level representations for linguistic, domain-specific and sentiment features—denoted \( E^{\texttt{(Lig)}}, E^{\texttt{(Dom)}}, E^{\texttt{(Sen)}} \), and their average \( E^{\text{(Avg)}} \) serves as the global feature representation.

When calculating the multi-head attention mechanism, attention is paid to the knowledge of the “Lig”, “Dom”, and “Sen” three levels. 
Unlike the traditional multi-head attention method that applies multi-head attention for “Lig”, “Dom”, and “Sen” respectively, we treat the “Lig”, “Dom”, and “Sen” as three heads respectively, as FaiMA mainly focuses on the knowledge of these three levels.

\subsubsection{Contrastive Learning}
Next, we will introduce the graph-level contrastive learning loss~\cite{10.1145/3534678.3539422}, which aims at optimizing the representation of the three aspects after the multi-head graph attention network.

In Section \ref{heuristic}, we obtain positive and negative sample pairs from linguistic, domain, and sentiment feature perspectives through heuristic rules. Similarly, taking the linguistic perspective as an example, for any sentence \( L_i \), we define its positive sample sentence set as \( \mathcal{P}_i \), and its negative sample sentence imposed as \( \mathcal{N}_i \). We take the global representation (sentence representation) obtained in Section \ref{network} as input to maximize the similarity between positive sample pairs and minimize the similarity between negative sample pairs. To this end, we define the contrastive learning formula as follows:
\begin{equation}  
\centering
    \mathcal{L}_{CL}^{\texttt{(Lig)}}=
     \frac{1}{B} \sum_{L_i} \left[-
    \log \frac{\sum_{L_j \in \mathcal{P}_i} \exp(\Gamma(E_i, E_j)/\tau)}{
   \sum_{L_k \in \mathcal{N}_i} \exp(\Gamma(E_i, E_k)/\tau)}
    \right] 
    \label{eq: CL Loss}
\end{equation}
where \( (L_i, L_j), L_j \in \mathcal{P}_i \) is the positive pair and \( (L_i, L_k), L_k \in \mathcal{N}_i \) is the negative pair for sentence \( L_i \). Moreover, we define the critic function as: \( \Gamma(u, v) = \cos (\text{Linear}(u), \text{Linear}(v)) \). \(\text{Linear}(\cdot)\) represents the projection function implemented with a two-layer perceptron model. \( \cos(\cdot) \) means cosine similarity, and we first normalize the embedding and then calculate point multiplication instead. \( \tau \in (0,1) \) is the annealing coefficient to avoid smoothing the exp function around 0 to speed up model convergence.

For linguistic features, domain features, sentiment features and global features, we calculate their respective contrastive learning loss functions \( \mathcal{L}_{CL}^{\text{(Lig)}}, \mathcal{L}_{CL}^{\text{(Dom)}}, \mathcal{L}_{CL}^{\text{(Sen)}}, \mathcal{L}_{CL}^{\text{(Avg)}} \) and sum them up as the model's contrastive learning loss:
\begin{equation}  
 \mathcal{L}_{CL} = \beta_1 \mathcal{L}_{CL}^{\text{(Lig)}}+  \beta_2 \mathcal{L}_{CL}^{\text{(Dom)}}+  \beta_3 \mathcal{L}_{CL}^{\text{(Sen)}} + \mathcal{L}_{CL}^{\text{(Avg)}},
\end{equation}
where \( \beta_1, \beta_2, \beta_3 \) are the reweighting coefficients. By continuously optimizing \( \mathcal{L}_{CL} \), we can achieve the goal of optimizing the model.

\subsection{Feature-aware Example Retrieval}
\label{Retriver}

After completing the MGATE training, the last phase of FaiMA involves example retrieval across different feature dimensions. Through the ICL method, the LLM can fully perceive the influence of different feature dimensions in the ABSA task on the output, thereby making more accurate predictions. Given an input sentence, the trained graph encoder \( \mathcal{G} \) can yield three feature vectors \( h_{\texttt{Lig}}, h_{\texttt{Dom}}, h_{\texttt{Sen}} \) and a pooled average vector \( h_{\texttt{Avg}} \). 
We only use the training set as the retrieval library and adopt the efficient FAISS (Facebook AI Similarity Search) algorithm\footnote{\url{github.com/facebookresearch/faiss}}~\cite{Johnson_Douze_Jegou_2021} for approximate nearest neighbor search:

\begin{equation} 
    \mathcal{N}_k(h) = \texttt{argmin}_{x_1, \ldots, x_k \in \text{S}} || h - x_i ||_2^2
\end{equation}
Here, \( k \) is the number of instances to be retrieved. For each feature dimension, we retrieve at least one nearest neighbour instance (i.e., \( k \geq 3 \)), and the retrieved instances are strictly de-duplicated.

For multi-domain ABSA tasks, we carefully design multiple different templates. The retrieved examples will be directly inserted into these templates to fine-tune the LLM in an ICL manner further.

\begin{table*}[h!]
  \centering
   \caption{Performances over five different runs with Macro-F1 score (\%) metric. The best performance is in bold and the second best results are underlined. } 
   \resizebox{1\linewidth}{!}{
  \begin{tabular}
  {c|ccccccccc|c}
    \toprule
    {}       & {laptop} & {restaurant} & {twitter} & {books} & {clothing} & {device} & {finance} & {hotel} & {service} & {Overall} \\
    \midrule
    BERT-CRF      & 55.32   & 68.15   & 57.85       & 42.15   & 61.41      & 55.78    & 55.37     & 61.44   & 54.77     & 58.03     \\
    SpanABSA-joint& 59.12   & 72.65   & 61.05       & 45.55   & 65.61      & 59.38    & 59.47     & 65.74   & 58.47     & 61.89     \\
    \midrule
    BART-Index    & 65.57   & 76.71   & 66.89       & 60.93   & 72.96      & 67.78    & 69.29     & 80.21   & 67.52     & 69.57     \\
    T5-Index      & 68.05   & 79.44   & 67.85       & 64.12   & \underline{78.59}      & \underline{71.13}    & 75.81     & 82.05   & 70.96     & 73.18     \\
    T5-Paraphrase & 68.29   & \underline{80.77}   & \textbf{69.52}       & 64.25   & \textbf{79.13}      & 70.82    & 75.77     & 82.17   & \underline{71.47}     & \underline{73.55}     \\
    \midrule
    LLaMA-SFT     & {68.50}    & {76.26}    & {62.80}        & {60.29}    & {73.71}       & \textbf{{71.21}}     & {73.74}      & \underline{82.87}    & {67.13}      & {70.54}      \\
    LLaMA-Random   & \underline{69.54}    & {78.23}    & {63.60}        & {64.07}    & {74.07}       & {66.82}     & \underline{76.56}      & {82.45}    & {71.29}      & {72.40}      \\
    LLaMA-SBERT   & {68.41}    & {76.80}    & {60.90}        & {62.32}    & {73.64}       & {67.58}     & {74.29}      & {81.04}    & {68.89}      & {70.99}      \\
    LLaMA-FaiMA   & \textbf{{70.58}}    & \textbf{{81.39}}    & \underline{68.00}        & \textbf{{66.33}}    & {77.08}       & {70.85}     & \textbf{{77.60}}      & \textbf{{83.45}}    & \textbf{{72.24}}      & \textbf{{75.62}}     \\
    \bottomrule
  \end{tabular}}
  \label{tab:main result}
\end{table*}

\section{Experiments}

\subsection{Dataset}
To bridge the absence of multi-domain benchmark datasets, we combine nine high-quality datasets from various domains into a comprehensive dataset, named MD-ASPE, including 14Restaurant \citelanguageresource{pontiki-etal-2014-semeval}, 14Laptop \citelanguageresource{pontiki-etal-2014-semeval}, Device \citelanguageresource{hu2004mining}, Service \citelanguageresource{toprak2010sentence}, Books, Clothing, Hotel \citelanguageresource{luo2022challenges}, Twitter \citelanguageresource{dong2014adaptive}, and Financial News Headlines \citelanguageresource{sinha2022sentfin}. MD-ASPE incorporates annotations from diverse teams and draws from rich data sources, effectively mimicking real-world multi-domain scenarios. We ensure data balance by employing random sampling strategies and standardizing the selected data by rectifying punctuation errors and addressing whitespace inconsistencies. Train and Test datasets statistics are summarized in Table \ref{tab:data}. 

\subsection{Baselines}

To rigorously and comprehensively evaluate our proposed approach, we chose a range of baseline models, from SLMs and Generative SLMs to LLMs. \underline{\textbf{1)}} Within the SLM category, we employ two cross-domain competent models based on the BERT framework \citep{devlin2018bert}: SpanABSA-joint (a span-level focused model) \citep{hu2019open} and BERT-CRF (a BERT-based model augmented with a CRF layer). \underline{\textbf{2)}} Generative SLMs include BART-Index based on BART \citep{raffel2020exploring}, its T5 variant T5-Index, and the variance model T5-Paraphrase (labels transduced into sequences using text templates) \citep{zhang2021aspect}. \underline{\textbf{3)}}  We also incorporate three LLM-based methods. The first is to conduct SFT directly based on LLaMA while keeping the instruction unchanged and only removing examples (LLaMA-SFT). Other ICL methods involved randomly selecting an equal number $k$ of examples (LLaMA-Random), and utilizing Sentence-BERT\footnote{\url{huggingface.co/sentence-transformers/all-MiniLM-L6-v2}} as sentence encoder to index and retrieve the most similar instances using the Euclidean algorithm (LLaMA-SBERT).

\subsection{Experimental Settings}

Our FaiMA comprises multiple stages, including the generation of pairs using heuristic rules, MGATE training, and SFT with ICL.
For the heuristic rules, we set $\theta_{Lig} = 0.43$, $\theta_{Dom} = 0.5$ and $\theta_{Sen} = 0.8$ to differentiate feature similarity.
Within the MGATE training phase, we employ the BERT-base-uncased\footnote{\url{huggingface.co/bert-base-uncased}} as a token encoder with an initial learning rate of $2 \times 10^{-4}$ running for 10 epochs.
In the ICL and SFT stage, we insert the 5 ($k=5$) most relevant examples into the prompt ordered by similarity \cite{liu2021makes}, including 2 average, 1 linguistic, 1 domain and 1 sentiment samples, respectively.
We use LLaMA2-7b\footnote{\url{huggingface.co/meta-llama/Llama-2-7b}} as the backbone model and leverage low-rank adaptation (LoRA) \cite{lora} for efficient parameter tuning, coupled with gradient accumulation and mixed-precision training. The learning rate and epochs are set to $8 \times 10^{-5}$ and 7, respectively. 
All methods use AdamW optimizer \cite{AdamW} with gradient decay, dynamic learning rate, and gradient clipping technique. 
The batch size $B$ is set to 128,  and $\tau=0.1, \delta=0.2, \beta_1=\beta_2=\beta_3=1$. 
All experiments are conducted on an Ubuntu 18.04.5 LTS server with an A800-80G GPU.
We randomly divide 10\% of the validation set from the training set, select the best-performing model on it, and employ the Macro-F1 value as the principal evaluation metric. 
We repeat experiments with different random seeds five times and report the average results.

\definecolor{mygreen}{RGB}{22,112,35}

\begin{table*}[]
    \centering
    \small
    \caption{Macro-F1 score of ablation experiment results on different datasets. Values in green indicate the drop in performance after removing a feature.}
    \setlength{\tabcolsep}{4pt}
    \resizebox{0.55\textwidth}{!}{%
    \begin{tabular}{lcccccc}
    \toprule
    Model & laptop & restaurant & twitter & books & clothing & avg. \\
    \midrule
    All & 70.58 & 81.39 & 68.00 & 66.33 & 77.08 & 73.94 \\
    w/o Lig. & \textcolor{mygreen}{\footnotesize(-1.35)} & \textcolor{mygreen}{\footnotesize(-2.10)} & \textcolor{mygreen}{\footnotesize(-0.95)} & \textcolor{mygreen}{\footnotesize(-1.80)} & \textcolor{mygreen}{\footnotesize(-1.22)} & \textcolor{mygreen}{\footnotesize(-1.75)} \\
    w/o Dom. & \textcolor{mygreen}{\footnotesize(-1.78)} & \textcolor{mygreen}{\footnotesize(-1.87)} & \textcolor{mygreen}{\footnotesize(-1.52)} & \textcolor{mygreen}{\footnotesize(-1.09)} & \textcolor{mygreen}{\footnotesize(-1.67)} & \textcolor{mygreen}{\footnotesize(-1.68)} \\
    w/o Sen. & \textcolor{mygreen}{\footnotesize(-0.37)} & \textcolor{mygreen}{\footnotesize(-0.92)} & \textcolor{mygreen}{\footnotesize(-0.68)} & \textcolor{mygreen}{\footnotesize(-0.53)} & \textcolor{mygreen}{\footnotesize(-0.86)} & \textcolor{mygreen}{\footnotesize(-0.71)} \\
    \bottomrule
    \end{tabular}
    }
    \label{tab:ablation}
\end{table*}

\subsection{Main Results}

Table \ref{tab:main result} shows the main experimental results. 
Our proposed LLaMA-FaiMA outperforms all baseline models in most domains, demonstrating an average performance gain of 3.22\% over the best previous method. This underscores the efficacy of the Feature-Aware ICL strategy in multi-domain ABSA scenarios. Among various SLMs, generative models such as BART, T5, and LLaMA evidently outclass BERT-based models. This superiority may stem from the generative models' more efficient pretraining methodology, which enables them to undergo large-scale unsupervised training on massive corpora, thereby acquiring richer and more diverse domain knowledge. 
Intriguingly, although the other LLaMA-based methods (LLaMA-SFT, LLaMA-Random, and LLaMA-SBERT) have larger model sizes than T5, their performance is somewhat lacking. 
We speculate that this could be due to the excessive size of LLM models, resulting in difficulties in learning transfer and adaptability. That efficient parameter fine-tuning alone may not be sufficient for optimal training \cite{Wang2023IsCA}. 
Despite employing a more advanced sentence encoder, for example, retrieval, LLaMA-SBERT experiences a decline in performance, indicating that conventional sentence encoding models struggle to adapt to the complexities of multi-domain ABSA tasks. 
In contrast, FaiMA provides stable examples from similar tasks, allowing the model to grasp the essence of the task at hand more rapidly. This demonstrates the effectiveness of our proposed approach and provides a robust new framework for the multi-domain ABSA task.

\subsection{Ablation Study}

To investigate the impact of different features on the performance of different domains, we sequentially remove three features (linguistic, domain, and sentiment) and then report the changes in the Macro-F1 score in the top five domains to validate the efficacy of the three features. The overall results are demonstrated in Table \ref{tab:ablation}.
Taking the results of the average drop, linguistic features have the most significant reduction to $-1.75$ in performance, followed by domain at $-1.68$ and sentiment feature at $-0.71$, substantiating the crucial role of linguistic features in ABSA tasks. 
Additionally, in some domains, such as Twitter, due to its unique characteristics, the impact of domain features is especially notable compared to linguistic features. 
In contrast, linguistic characteristics have the most significant impact in the restaurant and book domains.
Text data in linguistic domains are generally more structured, making them more susceptible to the influence of linguistic features. 
Meanwhile, the clothing and restaurant domains show a more pronounced dependency on sentiment features due to the high diversity in aspects and sentiments. 
The variation in the impact of linguistic features across domains is a reflection of unique language usage and contextual factors inherent to each domain.
Typically, the lack of any variation leads to a decrease in performance when compared to the complete model.

\subsection{Effectiveness Analysis of MGATE}

\begin{figure}[t]
\centering
\includegraphics[width=0.45\textwidth]{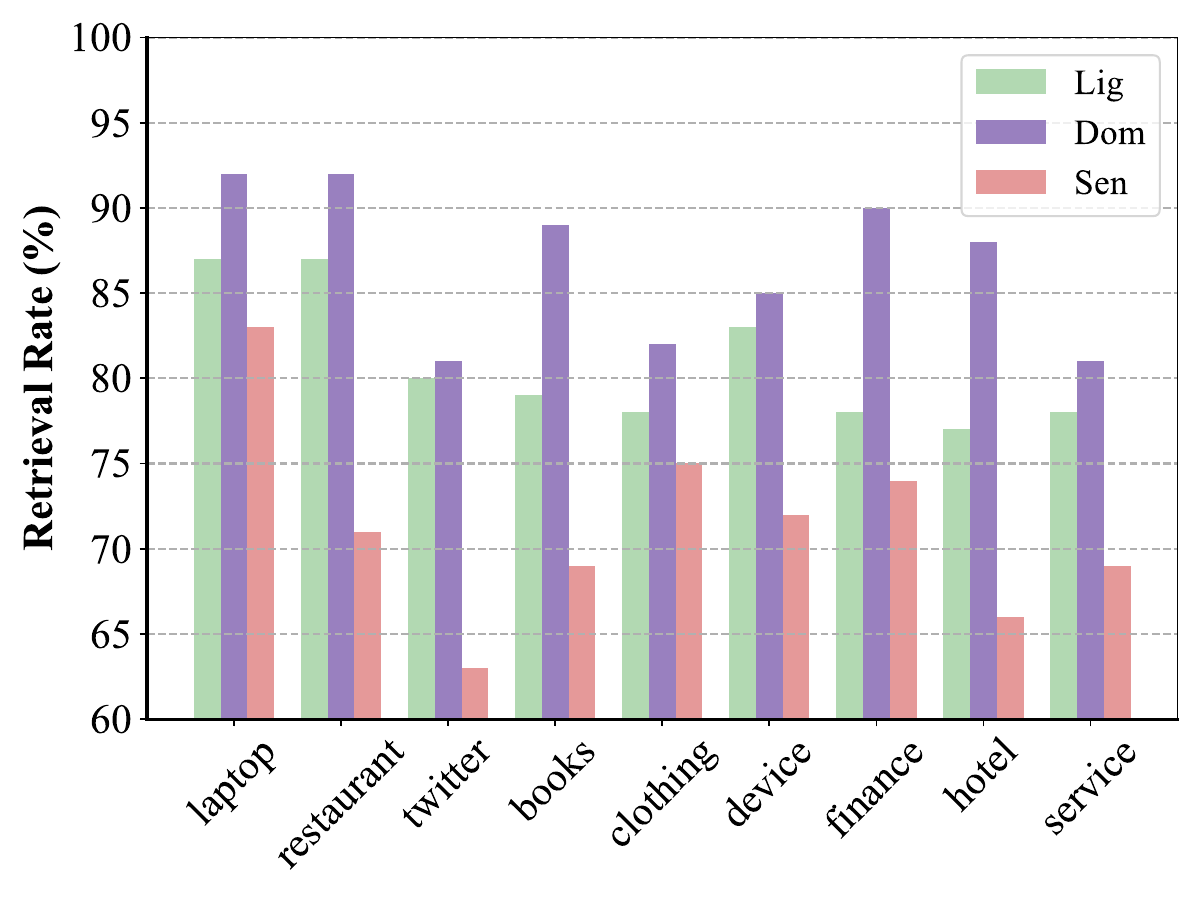}
\caption{The retrieval success rate of the three relevant feature examples retrieved for each domain.}
\label{fig:retriver_val}
\end{figure}

\begin{table*}[!h]
    \centering
    \small
    \setlength\tabcolsep{2pt}
    \caption{Case study reports two representative samples, including the retrieved most relevant examples on three features using MGATE and the prediction of FaiMA.}
    \label{tab:cases} 
    \begin{tabularx}{1.01\textwidth}{lXX}
        \toprule
        & \textbf{Case \#1} & \textbf{Case \#2} \\
        \midrule
        \textbf{Sample} & Input: The food was great - sushi was good, but the cooked food amazed us. \newline Output: [food, \redfont{positive}], [sushi, \redfont{positive}], [cooked food, \redfont{positive}] \newline Predict: [food, \redfont{positive}], [sushi, \redfont{positive}], [cooked food, \redfont{positive}] & Well, it happened because of a graceless manager and a rude bartender who had we waiting 20 minutes for drinks, and then tells us to chill out. \newline Output: [manager, \bluefont{negative}], [bartender, \bluefont{negative}], [drinks, \textcolor{mygreen}{neutral}] \newline Predict: [manager, \bluefont{negative}], [bartender, \bluefont{negative}]\\ \\ 
        
        \textbf{Lig.} & The service was excellent, the food was excellent, but the entire experience was very cool. \newline Output: [service, \redfont{positive}], [food, \redfont{positive}], [experience, \redfont{positive}] & The whole setup is truly unprofessional and I wish Cafe Noir would get some good stuff, because despite the current one this is a great place. \newline Output: [staff, \bluefont{negative}] \\ \\
        
        \textbf{Dom.} & The food was very expensive (we spent \$160 for lunch for two) but extremely tasty. \newline Output: [food, \redfont{positive}] & One would think we'd get an apology or complimentary drinks - instead, we got a snobby waiter who wouldn't even take our order for 15 minutes and gave us a lip when we asked him to do so. \newline Output: [waiter, \bluefont{positive}] \\ \\
        
        \textbf{Sen.} & The spicy tuna roll was unusually good and the rock shrimp tempura was awesome, great appetizer to share! \newline Output: [spicy tuna roll, \redfont{positive}], [rock shrimp tempura, \redfont{positive}], [appetizer, \redfont{positive}] & We actually gave 10\% tip (which we have never done despite mediocre food and service), because we felt totally ripped off. \newline Output: [food, \textcolor{mygreen}{neutral}] \\ 
        \bottomrule
    \end{tabularx}
\end{table*}

To validate the effect of MGATE (cf. Section \ref{MGATE}) for Feature-aware ICL components, we employ gpt-3.5-turbo\footnote{\url{openai/api/openai/chat-completion}} as an adjudicator to determine whether the examples retrieved by MGATE are similar in the validation set\footnote{Through testing, we found GPT can achieve human-like judgment due to excellent understanding ability.}. 
The retrieval rates for three features are illustrated in Figure \ref{fig:retriver_val} in various domains,
indicating that all three features achieve a relatively high success rate (over 50\%), proving the effectiveness of MGATE for multi-domain ABSA sentence encoding. 
Domain features exhibit the most explicit retrieval rate. 
Sentiment features are inferior to Linguistic features, and we attribute that Sentiment features are more multi-component and complex, leading to relatively low retrieval rates.

\subsection{Case Study}

To provide an insightful understanding of the efficacy of MGATE, we conduct case studies to detail retrieval and predictive results.
\underline{1)} For the correctly predicted Case \#1, we observe that all three examples show a high similarity to the input sentence in the corresponding feature dimensions. 
They share very similar syntactic structures linguistically, while also being from the same domain and possessing consistent sentiment polarity and quantity. 
These well-matched examples enable the model to fully apprehend each feature's role and effectiveness. 
\underline{2)} On the other hand, for Case \#2, the sentence composition and sentiment are somewhat complex, and only the domain feature successfully matching. 
The linguistic examples focus on partial similarity(``bartender,'' ``manager,'' and ``staff''), while the sentiment examples, possibly due to limited sample size, only offer support for the "neutral" label. 
Despite these limitations, the model still delivers accurate predictions, only overlooking the less frequent ``neutral'' label.

\section{Conclusion and Future Direction}
In this paper, we introduce FaiMA, a novel framework tailored to address the challenges of multi-domain Aspect-Based Sentiment Analysis (ABSA). 
The core insight of FaiMA is to utilize in-context learning (ICL) as a feature-aware tool in LLM. 
Moreover, FaiMA leverages GNNs and proposes MGATE, which captures the intricate interplay between linguistic, domain-specific, and sentiment features. Together with contrastive learning, MGATE empowers the model to retrieve highly analogous examples for any given input. 
Comprehensive experiments carried out in several domains demonstrate the effectiveness of FaiMA.

In summary, our research reveals the potential of LLMs in advancing ABSA studies, especially in multi-domain and cross-domain intricacies, providing a new insight and solution for integrating traditional GNN-based methods and LLMs, holding promise for broader sentiment analysis applications, i.e. Aspect Sentiment Triplet Extraction (ASTE) and Aspect Sentiment Quad Prediction (ASQP). 
Despite these successes, FaiMA, as an LLM-based model, needs higher training and deployment costs compared to previous methods. Another limitation of our model is its current focus on extracting binary sentiment elements, we plan to explore the extraction of triplet and quadruple and continue to build the appropriate dataset in future.

\section{Ethics Statement}

There are no ethics-related issues in this paper. We conduct experiments on publicly available datasets. These datasets do not share personal information and do not contain sensitive content that can be harmful to any individual or community.


\section{Acknowledgments}

The authors thank anonymous reviewers for their insightful comments. This work is mainly supported by the Key Program of the Natural Science Foundation of China (NSFC) (Grant No. U23A20316) and Key R\&D Project of Hubei Province (Grant No.2021BAA029).


\nocite{*}
\section{Bibliographical References}\label{sec:reference}

\bibliographystyle{lrec-coling2024-natbib}
\bibliography{lrec-coling2024-example}

\section{Language Resource References}
\label{lr:ref}
\bibliographystylelanguageresource{lrec-coling2024-natbib}
\bibliographylanguageresource{languageresource}

\end{document}